\documentclass[sigconf]{acmart}
\usepackage{subfig}
\usepackage{booktabs}
\usepackage{enumitem}
\usepackage{multirow}
\usepackage{tabularx}
\usepackage{caption}
\usepackage{subcaption}
\usepackage{float}
\usepackage{listings}
\renewcommand{\lstlistingname}{Prompt}

\AtBeginDocument{%
  \providecommand\BibTeX{{%
    \normalfont B\kern-0.5em{\scshape i\kern-0.25em b}\kern-0.8em\TeX}}}
\usepackage{fancyhdr}
\renewcommand\footnotetextcopyrightpermission[1]{} 
\setcopyright{none}

\begin{document}

\title{Language-Guided Self-Supervised Video Summarization Using Text Semantic Matching Considering the Diversity of the Video}

\author{Tomoya Sugihara}
\affiliation{%
  \institution{The University of Tokyo}
  \city{Tokyo}
  \country{Japan}}
\email{sugihara@cvm.t.u-tokyo.ac.jp}

\author{Shuntaro Masuda}
\affiliation{%
  \institution{The University of Tokyo}
  \city{Tokyo}
  \country{Japan}}
\email{masuda@cvm.t.u-tokyo.ac.jp}

\author{Ling Xiao}
\affiliation{%
  \institution{The University of Tokyo}
  \city{Tokyo}
  \country{Japan}}
\email{ling@cvm.t.u-tokyo.ac.jp}

\author{Toshihiko Yamasaki}
\affiliation{%
  \institution{The University of Tokyo}
  \city{Tokyo}
  \country{Japan}}
\email{yamasaki@cvm.t.u-tokyo.ac.jp}




\renewcommand{\shortauthors}{Sugihara et al.}
\begin{abstract}
Current video summarization methods rely heavily on supervised computer vision techniques, which demands time-consuming and subjective manual annotations. To overcome these limitations, we investigated self-supervised video summarization. Inspired by the success of Large Language Models (LLMs), we explored the feasibility in transforming the video summarization task into a Natural Language Processing (NLP) task. By leveraging the advantages of LLMs in context understanding, we aim to enhance the effectiveness of self-supervised video summarization. Our method begins by generating captions for individual video frames, which are then synthesized into text summaries by LLMs. 
Subsequently, we measure semantic distance between the captions and the text summary.
Notably, we propose a novel loss function to optimize our model according to the diversity of the video.
Finally, the summarized video can be generated by selecting the frames with captions similar to the text summary. Our method achieves state-of-the-art performance on the SumMe dataset in rank correlation coefficients.
In addition, our method has a novel feature of being able to achieve personalized summarization.
\end{abstract}




\maketitle

\section{Introduction}
\begin{figure*}[t]
    \centering
    \includegraphics[width=1.0\linewidth]{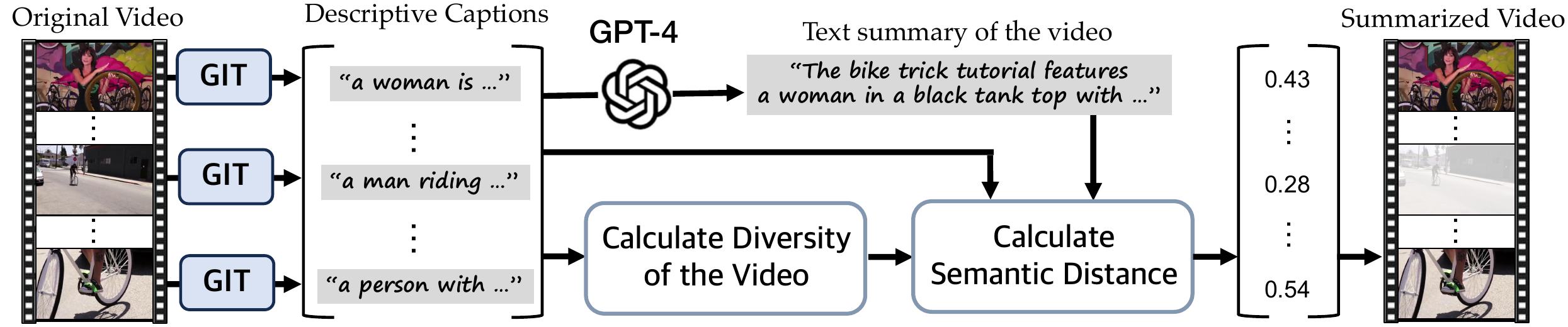}
    \caption{Overview of our proposed framework. We take only videos as input and first generate captions from individual frames using a pre-trained image captioning model, the Generative Image-to-text Transformer (GIT)~\cite{git}. The text summary is then created by GPT-4~\cite{gpt4}. The semantic distance between individual captions and the text summary is calculated using the proposed Preserving Diversity Loss (PDL) to calculate frame-level scores. Finally, the frame-level scores are aggregated into scene-level scores, and the knapsack problem is solved to select a subset of scenes, thereby creating the video summary.}
    \label{fig:overview}
\end{figure*}

Video summarization involves distilling a full-length video into a concise version that encapsulates the most crucial or engaging elements of the original. The goal is to produce a summary that is brief yet delivers a cohesive grasp of the principal themes or narratives of the video. Video summarization has emerged as an important research topic in today's fast-paced information society for two main reasons:~1) There has been an unprecedented increase in video content across social media platforms. This includes not only professional productions such as news broadcasts, live concerts, and sports events but also user-generated content. Dominant platforms especially YouTube and Instagram have become integral to various facets of our daily lives, and they are expected to maintain their far-reaching impact\footnote{\url{https://www.statista.com/statistics/1061017/digital-video-viewers-number-worldwide/}}.~2) The overwhelming volume of available video content, coupled with the modern demand for rapid assimilation of extensive information, underscores the growing necessity for video summarization technology. As it becoming increasingly time-consuming for individuals to consume and process all the available material, video summarization is proving to be an essential development within multimedia and computer vision to address these societal demands.

Video summarization, which creates an abridged version of a video while preserving its essential content and information, has wide-ranging applications. It enables users to quickly absorb the crucial parts of lengthy videos, thus optimizing the time spent understanding the content. For example, in educational contexts, summarization can boost learning efficiency by concentrating on key topics. It also serves to highlight the most thrilling or significant moments, thereby improving the viewer's experience. Moreover, the technology is instrumental for producing promotional or commercial clips, showcasing its versatility across different domains.


However, video summarization is inherently complex due to the diverse content and subjectivity involved in identifying key segments within extensive footage. Sophisticated analytical approaches are required to discern these crucial shots. With the evolution of deep neural network architectures, the accuracy of computer vision-based approach has significantly improved.
Techniques such as Convolutional Neural Networks (CNNs) for image analysis~\cite{vsum,sparsity,global,MSVA}, Recurrent Neural Networks (RNNs) for temporal sequence modeling~\cite{lstm,graph,logcosh}, and the attention mechanisms for highlighting important features~\cite{logcosh,CASUM,RSSUM} have enabled effective summarization models.
Numerous studies have pursued supervised learning approaches in video summarization, seeking models of heightened accuracy~\cite{a2summ,SSPVS,MSVA,scale,MTIDNet,clipit,mtap,iptnet,csta}.
However, these supervised video summarization methods rely heavily on large amounts of human-generated annotated data, which is time-consuming and subjective. As a result, low-quality labeled data can significantly restrict the performance.

Conversely, the latest progress in Large Language Models (LLMs), including the Generative Pre-trained Transformer 4 (GPT-4)~\cite{gpt4} and Large Language Model Meta AI 2 (LLaMA 2)~\cite{llama2}, has greatly advanced text summarization, enabling the generation of accurate summaries in zero-shot scenarios. This breakthrough in LLMs has opened up new possibilities to video summarization, offering potential solutions to the challenges associated with supervised methods. 
Furthermore, due to the advancements in Vision-and-Language models, integrating vision and language has become more straightforward. This progress has simplified tasks such as Visual Question Answering (VQA)~\cite{vqa}, image-to-text~\cite{git} or text-to-image task~\cite{stable}. Image captioning models, which produce descriptive captions for images, represent a captivating convergence of computer vision and natural language processing (NLP), and are instrumental for tasks that translate visual content into text.

To address the aforementioned issues in the video summarization task, this paper explores the potential of LLMs in video summarization, which have demonstrated effectiveness in natural language processing and contextual understanding~\cite{gpt3mix,generatingdata,clipIQA,gpt4,llama2,videorepresentation}. To this end, we propose a novel self-supervised method that effectively leverages LLMs for video summarization and introduce a loss function tailored to the video's diversity, thereby fostering the development of a robust video summarization model. We term this proposed loss function the Preserving Diversity Loss (PDL).
Briefly, our contributions are as follows:
\begin{itemize}
      \item We are the first to transform a video summarization task into a semantic textual similarity task in a self-supervised way.
      \item We propose a novel self-supervised video summarization method that leverages LLMs, unlocking their potential in natural language and contextual/semantic understanding to enhance video summarization.
      \item  We mathematically analyze the characteristics of the dataset and adjust the balancing parameter based on the diversity of the video to further enhance our model. Experimental results verified the effectiveness of our proposed model. 
      \item  Our proposed framework is versatile and can incorporate user specifications to generate user-guided video summaries, which has been very difficult in previous approaches.
\end{itemize}

\section{Related Work}
\subsection{Video Summarization}
Video summarization involves processing a sequence of video frames and producing a binary vector that indicates which shots to be included in the summary.
Conventional video summarization methods can be categorized into unsupervised~\cite{vsum,sparsity,global,lstm,graph,logcosh,CASUM,RSSUM,M3SUM} and supervised learning methods~\cite{a2summ,SSPVS,MSVA,scale,MTIDNet,clipit,mtap,iptnet,csta}.
Unsupervised video summarization methods calculate frame-level scores by using frame images without relying on annotated data. Recent unsupervised video summarization methods employ Generative Adversarial Network (GAN) based methods~\cite{sparsity,global,csnet}, RNNs based methods~\cite{lstm,graph,logcosh} and self-attention module based methods~\cite{logcosh,CASUM,RSSUM,csnet} to calculate frame-level scores. 
SAM-GAN~\cite{sparsity} uses GAN to select a subset of keyframes, aiming to generate summaries that closely resemble the original video. DSAVS~\cite{logcosh} calculates the similarity between the caption and frame images within the same semantic space, employing Long Short Term Memory (LSTM) module and a self-attention module. RSSUM~\cite{RSSUM} adopts self-supervised learning by training an encoder to reconstruct missing video sections using rule-based masked operations.

Supervised learning for video summarization involves methods that use large-scale, manually annotated frame-level data for training~\cite{a2summ,SSPVS,MSVA,scale,MTIDNet,clipit,mtap,iptnet,csta}.
MSVA~\cite{MSVA} extracts various visual features from frame images, such as static and dynamic features. A2Summ~\cite{a2summ} aligns and attends the multimodal input by an alignment-guided self-attention module to make the use of cross-modal correlation.
In SSPVS~\cite{SSPVS}, self-supervised learning is conducted on both the text encoder and the image encoder during pre-training, with the training data being used for multi-stage fine-tuning in downstream tasks.
However, constructing manually annotated datasets is not only time-consuming and expensive, but also challenging. This process necessitates reducing subjectivity among annotators. Therefore, creating training data for video summarization is not a long-term solution~\cite{rethinking,mrhisum,annotation}. It is essential to develop self-supervised video summarization technologies for real-world applications.

\subsection{Text Summarization}
Text summarization can be divided into two types: extractive~\cite{matchsum,sentencecontra,emotion,lapata} and abstractive summarization~\cite{llama2,gpt4,gemini,bart}. Extractive summarization entails selecting significant sentences or phrases from a document and combining them to create a text summary. 
MatchSum~\cite{matchsum} formulates this as a semantic text matching problem, where the correct summary is semantically embedded closer to the original document compared to other candidate summaries. Specifically, the model employs a Siamese-BERT architecture that is based on the Siamese network~\cite{siamese}. The Siamese-BERT consists of two Bidirectional Encoder Representations from Transformers (BERT)~\cite{bert} with shared-weights and a cosine similarity layer.

Abstractive summarization involves concisely paraphrasing the main content from the sentences within the input document while retaining the important parts. Recent advancements have been marked by the development of LLMs such as GPT-4~\cite{gpt4} and LLaMA 2~\cite{llama2}, which are built upon the Transformer~\cite{transformer} architecture and have been pre-trained on a vast amount of datasets. These models possess advanced language understanding and generation capabilities, improving the ability to accurately summarize long documents. Also, it is reported that these performances are equivalent to human-written summaries~\cite{llmsurvey,evaluateLLM}. 
As a result, text summarization technology has become capable of accurately summarizing more complex and lengthy texts in zero-shot settings.

\section{Method}
\subsection{Framework of our method}
\begin{figure}[t]
    \centering
    \includegraphics[width=1.0\linewidth]{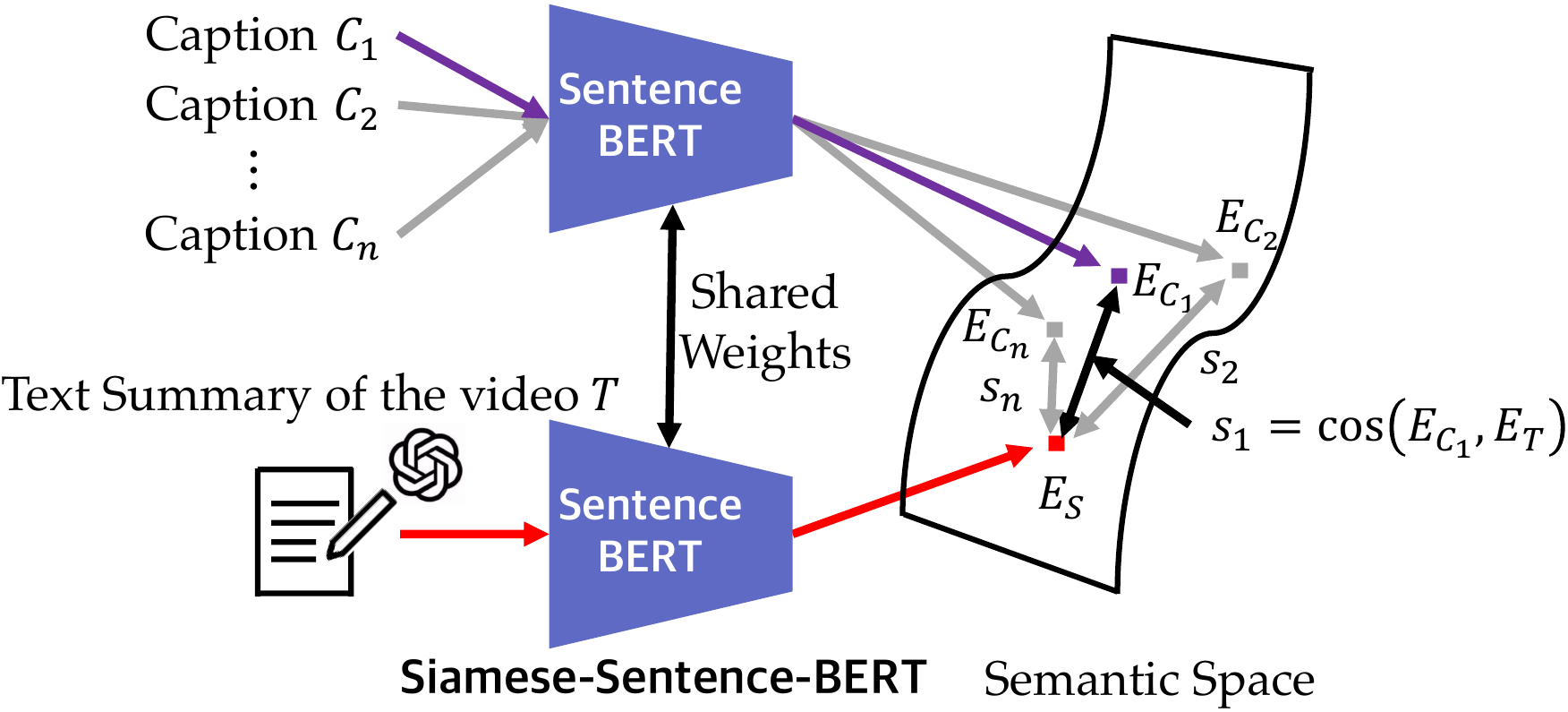}
    \caption{The pipeline of the proposed semantic distance calculation module. We solve semantic textual similarity task between individual frame captions and the generated text summary to calculate frame-level scores using Siamese-Sentence-BERT architecture~\cite{siamese,sentence}.}
    \label{fig:model}
\end{figure}

As mentioned before, our motivation is to leverage LLMs for unsupervised video summarization, freeing the process from the burden of extensive data annotation and the subjective errors associated with it. 
We transform the video summarization task into an NLP task, enabling videos to be represented linguistically and fully taking advantage of LLMs in contextual/semantic understanding. 
Figure~\ref{fig:overview} provides the overview of our framework. Our method begins by generating descriptive captions for downsampled individual video frames by using a pre-trained image captioning model, Generative Image-to-text Transformer (GIT)~\cite{git}. Then, these captions are synthesized into a coherent text summary by using GPT-4~\cite{gpt4}. Afterwards, the number of downsampled captions is denoted as $n$, with $C_{i}$ representing the frame captions for $i \in \{1, \dots ,n\}$, and $T$ representing the generated text summary. We begin with encoding each of the $n$ frame captions and the generated text summary using pre-trained Sentence-BERT~\cite{sentence} as shown in Figure~\ref{fig:model}. The encoded captions are denoted as $E_{C_{i}}$, and the encoded text summary as $E_T$. Then, we perform deep metric learning on the Siamese-Sentence-BERT architecture~\cite{sentence,siamese}, which consists of two pre-trained Sentence-BERTs with shared-weights and a cosine similarity layer. 
After the model is trained, the model inputs the pair of individual captions, $C_i$ and the text summary, $T$. The encoded captions are denoted as $E'_{C_{i}}$ and the encoded text summary is denoted as $E'_{T}$.
The similarity score $s_i$ between $E'_{C_i}$ and $E'_T$ is calculated using the cosine similarity, calculated by Eq.~(\ref{eq:s_i}) as follows:
\begin{equation}
s_i = \frac{{E'_{C_i}}^T E'_T}{\|E'_{C_i}\|\|E'_T\|} (= \cos(E'_{C_i},E'_T)).
\label{eq:s_i}
\end{equation}
Then, $i$-th frame-level score $S_i$ is calculated as follows:
\begin{equation}
S_i = \frac{1}{2}(1+s_i). 
\label{norm}
\end{equation}
In terms of computational complexity, we calculate the semantic textual similarity for individual captions and the text summary separately, resulting in a complexity of $\mathcal{O}(d \cdot n)$, where $d$ is the hidden dimension.

\subsection{Loss function}
The loss function of our model consists of two components, an improved margin ranking loss and a sparsity regularization loss~\cite{sparsity}). Details of these two components will be given below.

\noindent\textbf{Improved Margin Ranking Loss.}
The conventional margin ranking loss is used for learning the relative relevance between different items. Normally, it has positive and negative pairs as input. In our task, we only care about the absolute relevance. Therefore, we propose an improved margin ranking loss tailored for the video summarization task.
It aims to increase the score difference between captions and the text summary, which is defined as:
\begin{equation}
\mathcal{L}_{m} = \frac{1}{n}\sum_{i=1}^n \max(0, -\lvert   S_i - S_{\rm{avg}} \rvert + m),
\label{eq:L_formula}
\end{equation}
where $S_{\rm {avg}}$ is the average value of $S_i$ and is calculated as follows:
\begin{equation}
S_{\rm {avg}}=\frac{1}{n}\sum_{i=1}^n S_i.
\label{avg}
\end{equation}  
Also, $m$ is the minimum score difference the model should maintain between the positive and negative classes. We classify $S_i$ into two classes according to the following rules. It is considered a score belonging to the positive class if $S_i$ is greater than $S_{\rm {avg}}$, and conversely, it is considered a score belonging to the negative class. Therefore, the larger the $m$ is, the greater the difference in scores between the positive and negative classes when using the model after training would become. 

\noindent
\textbf{Regularization Loss.}
We apply this sparsity regularization loss to ensure a more concise and informative summary as exiting works~\cite{CASUM,logcosh,vsum,sparsity,weak}, which is defined as:
\begin{equation}
    \mathcal{L}_{{s}} = \left\| \frac{1}{n} \sum_{i=1}^{n} S_i - \epsilon \right\|_2,
\label{regular}
\end{equation}
where $\epsilon$ is a hyperparameter that specifies the proportion of frames to be selected as key frames and set to 0.3 as in existing works~\cite{logcosh,sparsity,weak}.
By constraining the number of key frames, summaries become more concise and relevant, avoiding redundancy and focusing on the most informative parts of the video.

\subsection{Impact of the loss functions}
\begin{figure*}[t]
\centering
\begin{minipage}[b]{0.65\columnwidth}
    \centering
    \includegraphics[width=1\columnwidth]{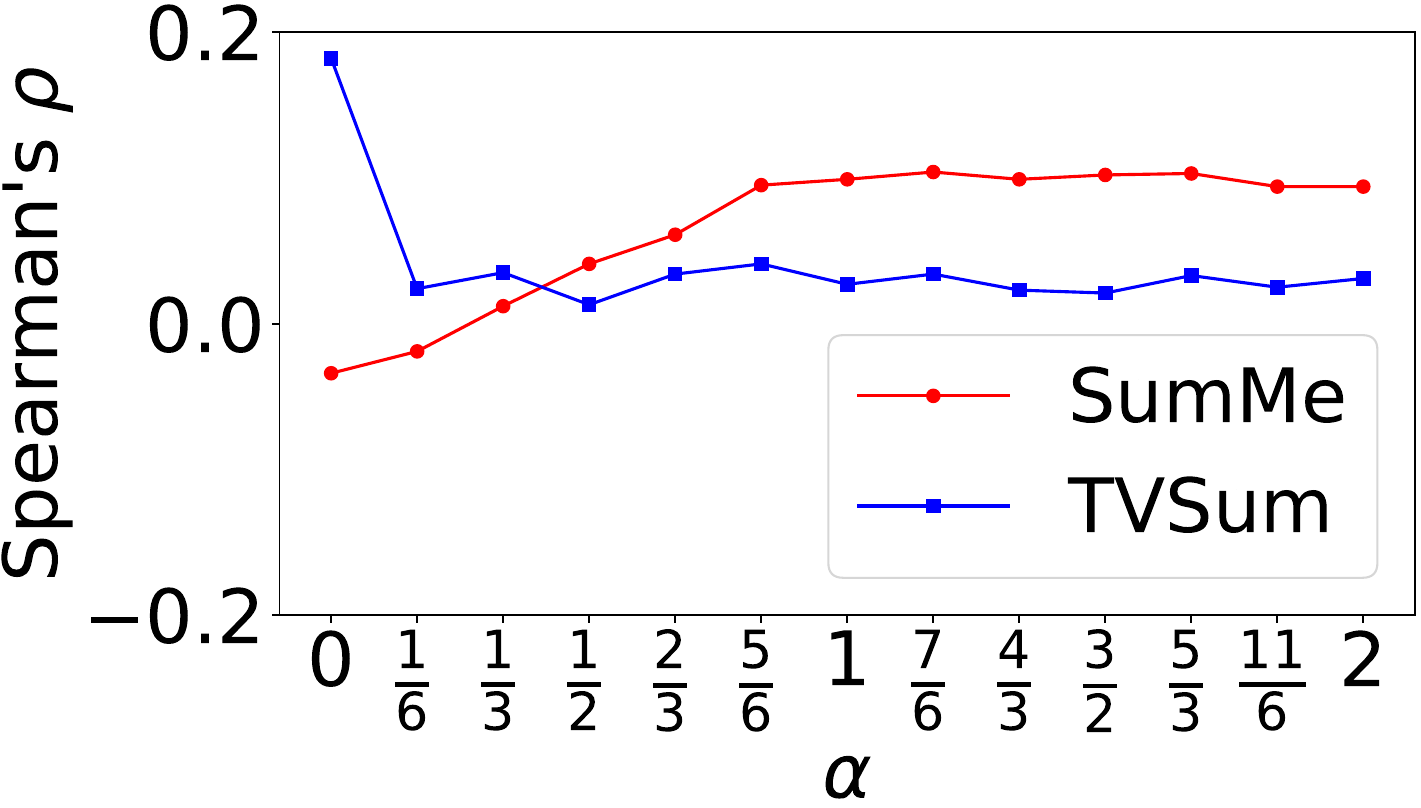}
    \caption{Effect of the fixed contribution value ($\alpha$) of the regularization loss~\cite{sparsity}. A higher value of Spearman's $\rho$ indicates a better video summary. }
    \label{fig:rho}
\end{minipage}
\hspace{5pt} 
\begin{minipage}[b]{1.3\columnwidth}
    \centering
	\subfloat[SumMe]{\includegraphics[width=0.43\textwidth]{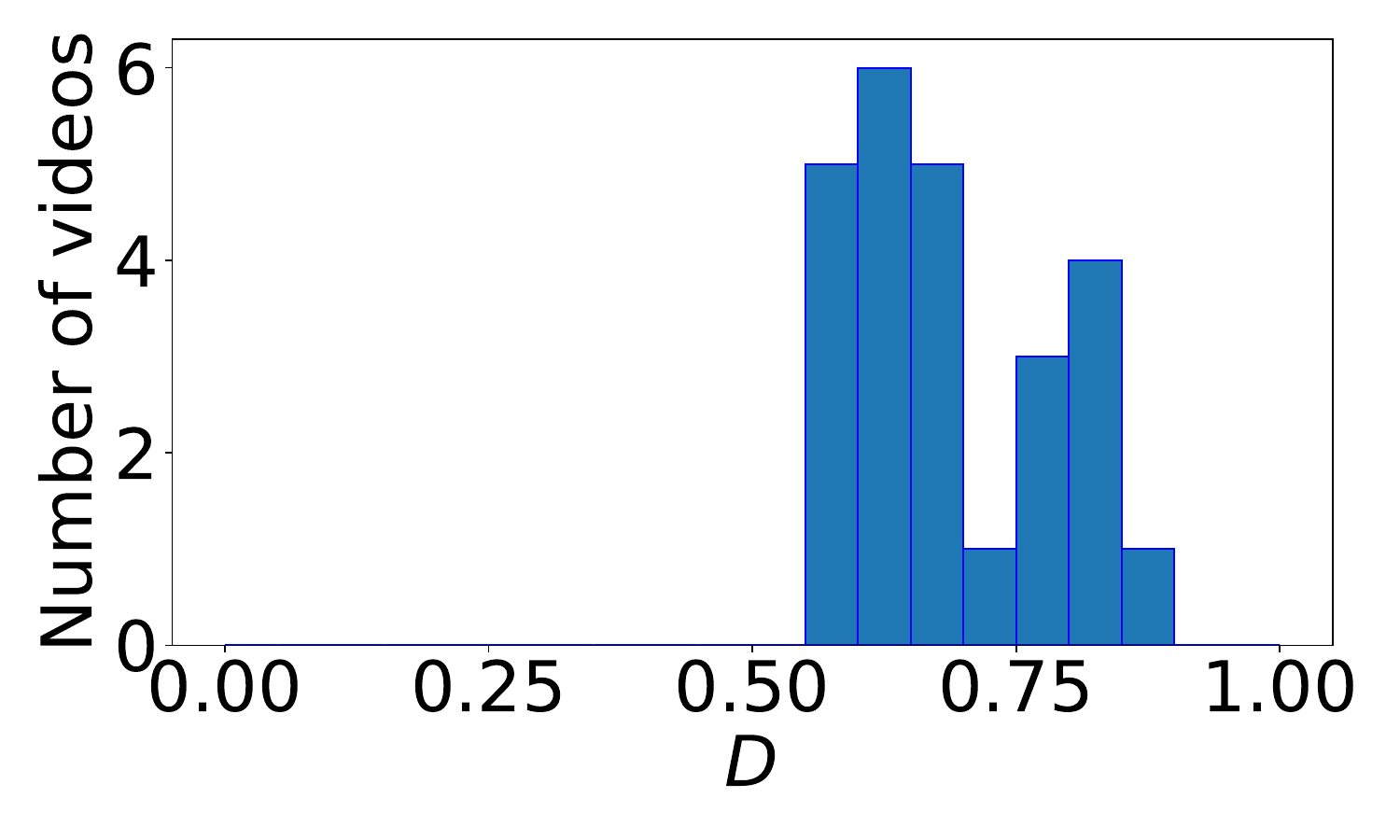}}\qquad
	\subfloat[TVSum]{\includegraphics[width=0.43\textwidth]{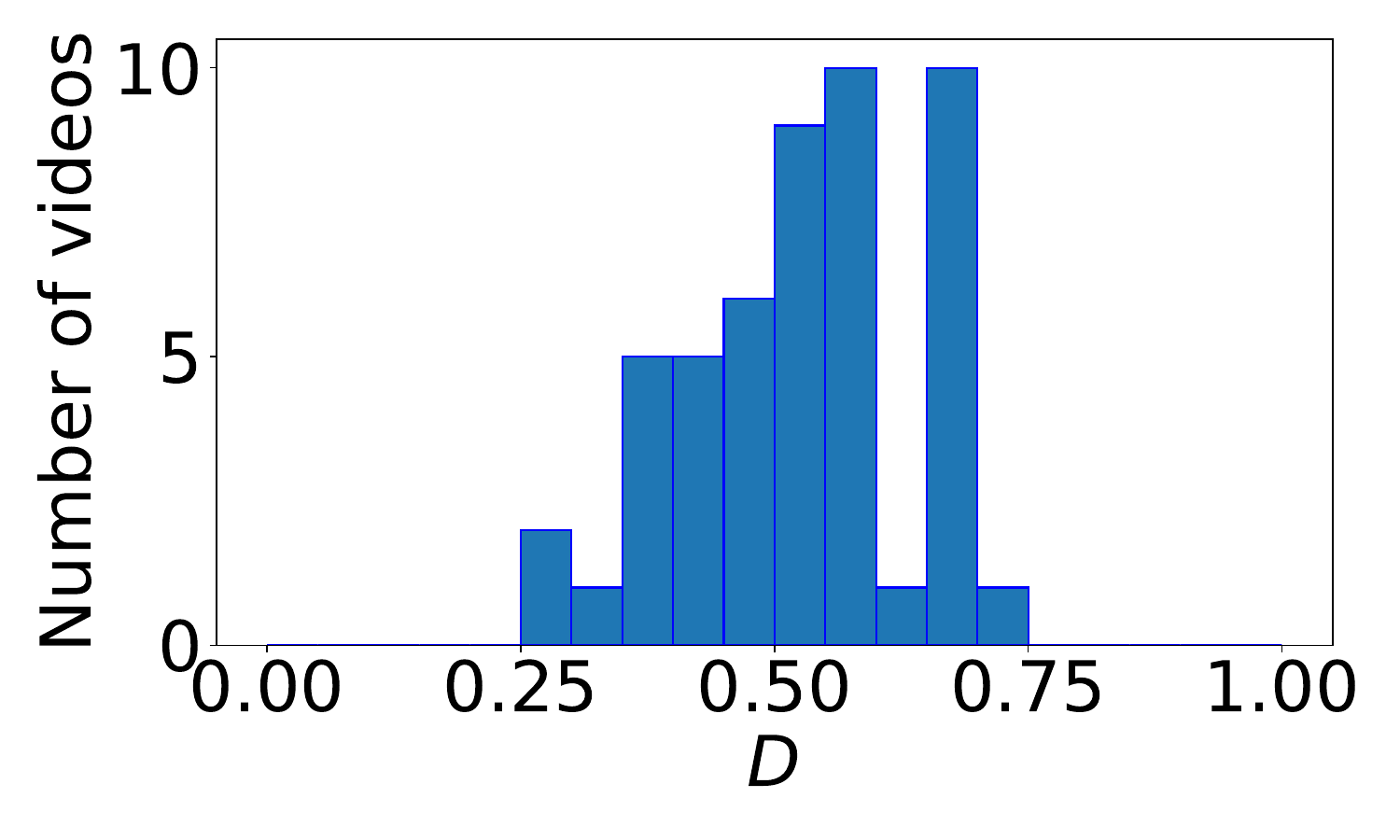}}\\
	\caption{The histogram of $D$ for individual videos within the SumMe~\cite{SumMe} and TVSum~\cite{TVSum} datasets. A lower value indicates that the linguistically represented video has lower diversity. TVSum has more videos with higher $D$ scores compared to SumMe.}
	\label{fig:dataset_diversity}
\end{minipage}
\end{figure*}
The improved margin ranking loss focuses on enlarging score difference between frame captions and the text summary,
and the regularization loss aims at control the sparsity of the selection vector~\cite{sparsity}. We investigated the contributions of the regularization loss. 
We conducted experiments, assigning a fixed value ($\alpha$) to the regularization loss. The overall loss was constructed as follows:
\begin{equation}
    \mathcal{L} = \mathcal{L}_{m} + \alpha \mathcal{L}_{s}.
\label{eq:alpha}
\end{equation}
Figure~\ref{fig:rho} shows the quality of the generated video summaries when varying $\alpha$ using two video summarization datasets: SumMe~\cite{SumMe}
and TVSum~\cite{TVSum}. It shows that a larger $\alpha$ value in SumMe results in higher quality video summaries, while in TVSum, a smaller $\alpha$ produces better quality.

\subsection{Proposed PDL}
\begin{figure*}[t]
	\centering
	\subfloat[Video 21 in SumMe (``Uncut Evening Flight'')]{\includegraphics[width=0.38\textwidth]{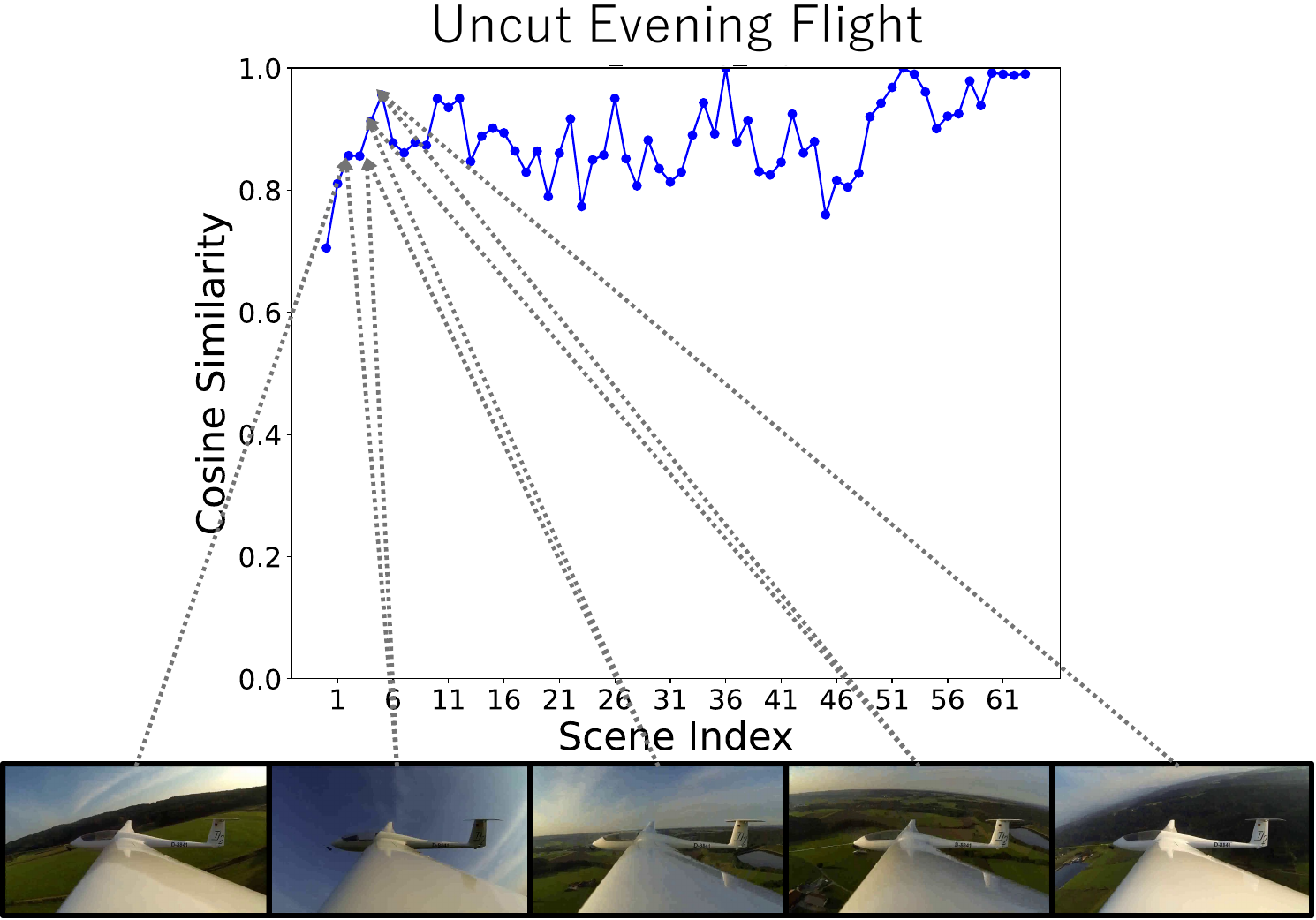}}\qquad
	\subfloat[Video 43 in TVSum (``How to stop your Fixie'')]{\includegraphics[width=0.38\textwidth]{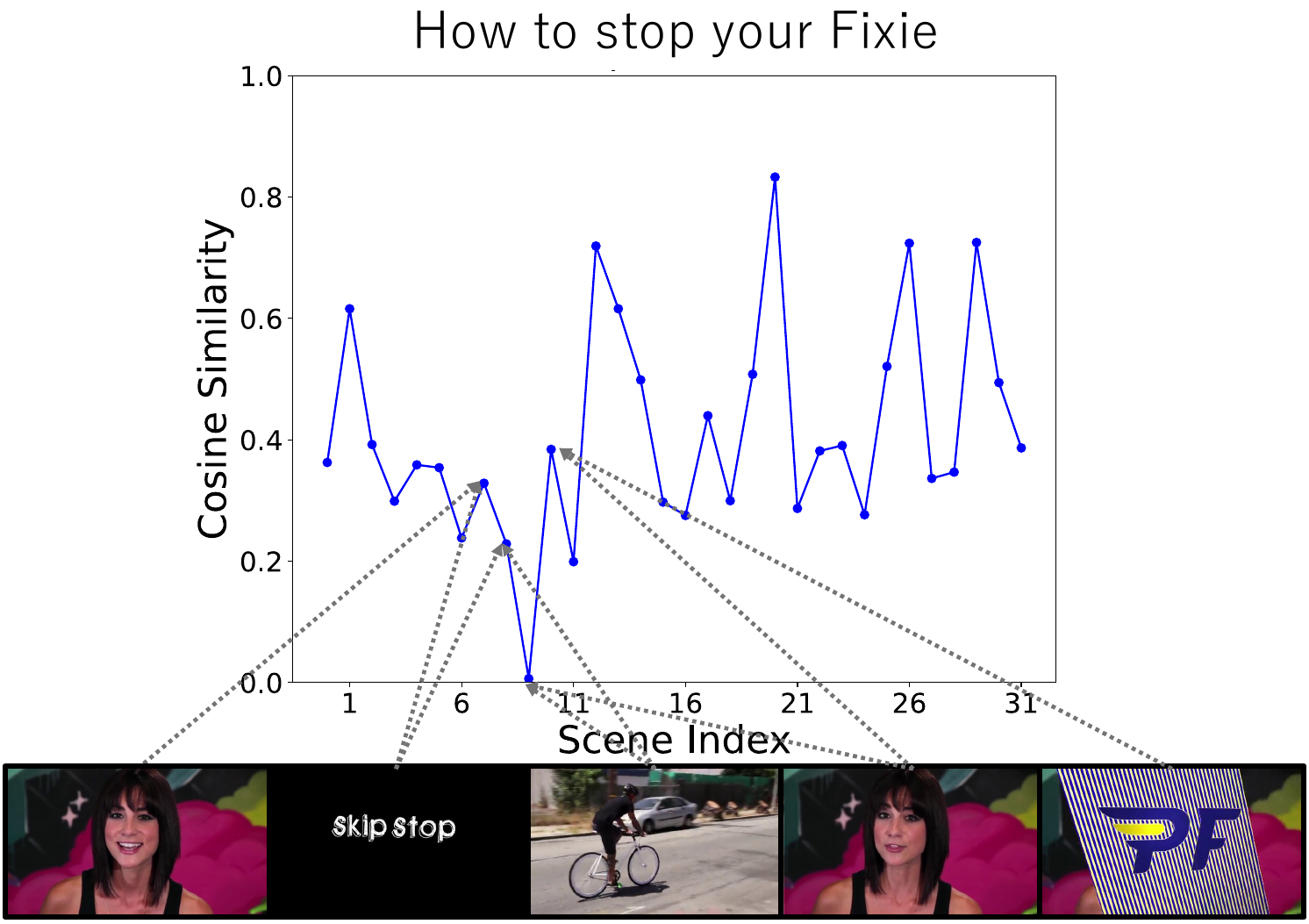}}\\
	\caption{Examples of the transition of $s_{\rm{change}_k}$. The cosine similarity between adjacent scenes are shown. A lower value indicates a reduced similarity to the linguistically represented adjacent scene.}
	\label{fig:example_diversity}
\end{figure*}
Based on the aforementioned observations, the $\alpha$ value should be defined differently to account for the varying characteristics of different datasets. An adaptive loss function is required to effectively handle general video summarization tasks. We also find that these differences are fundamentally linked to the diversity of the videos within each dataset. Therefore, we investigate the diversity of the datasets.
Firstly, based on encoded captions generated from all the video frames, we employed Kernel Temporal Segmentation (KTS)~\cite{kts}, an algorithm used in conventional video summarization~\cite{vsum,graph,logcosh,CASUM,RSSUM} for segmenting frames into scenes, to segment the linguistically represented video into scenes of similar content. 
The number of scenes is denoted as $q$, and the averaged feature value of the encoded captions generated from all the video frames within the same $j$-th scene is represented as $E_{\rm{scene}_j}$.
When the $j$-th scene consists of $p$ video frames, $E_{\rm{scene}_j}$ is calculated as follows:
\begin{equation}
    E_{\rm{scene}_j} = \frac{1}{p} \sum_{i=1}^{p} E_{C_i}.
\end{equation}
When the cosine similarity between $j$-th and ($j+1$)-th scenes is $k$-th adjacent scene, we define it as $s_{\rm{change}_k}$ and calculated as follows: 
\begin{equation}
    s_{\rm{change}_k} = \sum_{j=1}^{q-1} \cos(E_{\rm{scene}_j}, E_{\rm{scene}_{j+1}}).
\label{eq:diversity}
\end{equation}
Then, the similarity between the linguistically represented video frames is defined as ${\rm{sim}}_{\rm{scene}}$, and $D$ is defined as the diversity of the video, calculated as follows:
\begin{equation}
    {\rm sim}_{{\rm scene}} = \frac{1}{q-1} \sum_{j=1}^{q-1} s_{\rm{change}_k},\quad
    D = 1 - {\rm sim}_{{\rm scene}}.
\label{eq:diversity}
\end{equation}
Note that a higher ${\rm sim}_{{\rm scene}}$ results in a lower $D$ score, indicating that when the similarity between adjacent scenes is high, the diversity of the video is low.
Figure~\ref{fig:dataset_diversity} shows the distribution of $D$ in videos within SumMe and TVSum. It shows that videos in SumMe have lower $D$ scores, suggesting the presence of many semantically similar scenes. Conversely, videos in TVSum have relatively higher $D$ scores, indicating the diversity of each video in the dataset is high.
Also, Figure~\ref{fig:example_diversity} shows examples of the transition of $s_{\rm{change}_k}$. In adjacent scenes, $s_{\rm{change}_k}$ is higher for linguistically similar videos. In contrast, linguistically different videos have lower $s_{\rm{change}_k}$ .

Building upon the aforementioned findings, we introduce a novel PDL, denoted as $\mathcal{L}_{\rm PDL}$. It adapts the contribution of the regularization loss based on video diversity and can effectively function across different video domains.
The mathematical definition of $\mathcal{L}_{\rm PDL}$ is
\begin{equation}
    \mathcal{L}_{\rm PDL} = \mathcal{L}_{m} + \lambda \mathcal{L}_{s},
\label{eq:overall}
\end{equation}
where $\lambda$ is an adaptive value to dynamically adjust the contribution of $\mathcal{L}_{s}$, determined by the video diversity and defined as follows:
\begin{equation}
\lambda = 
\begin{cases} 
0 & \text{if } D \geq  \delta, \\
(1- D) \exp (1 - D) & \text{if } D < \delta,
\end{cases}
\label{div}
\end{equation}
where $\delta$ denotes a threshold to measure the diversity of the video. 
The values of 0 and $(1- D) \exp (1 - D)$ 
are decided empirically. 
When the video diversity is high, incorporating the regularization loss becomes unnecessary, as inherent diversity ensures a rich and comprehensive representation of the content. Conversely, if similar captions are prevalent, making it challenging to differentiate based on similarity, it is necessary to reduce the contribution of the improved margin ranking loss.

\section{Experiments}

\subsection{Datasets}
We evaluate our method on two standard video summarization datasets, SumMe~\cite{SumMe} and TVSum~\cite{TVSum}, to compare it with previous works~\cite{vsum,csnet,graph,logcosh,CASUM,RSSUM}.

\noindent\textbf{SumMe}  comprises 25 unedited personal YouTube videos that capture various events, such as cooking and sports, with each video ranging in length from 30 seconds to 6 minutes. The title of the video is available as metadata. Annotations were created over a total of more than 40 hours by 15 to 18 annotators, with the audio track not included.

\noindent\textbf{TVSum} contains 50 edited YouTube videos, spanning 10 categories including dog shows and parades, with 5 videos from each category. Each video lasts between 1 to 10 minutes. Its metadata encompasses titles, genres, and query categories. The annotations are provided by 20 annotators, who watched the videos without audio.

\subsection{Implementation Details}
We first downsample the input video to two frames per second as existing works~\cite{vsum,csnet,graph,logcosh,CASUM,RSSUM}. Then, we process the downsampled frames with pre-trained GIT~\cite{git} using ``a photo of'' as a prompt and the prompt is excluded when generating the text summary. 
For the text summaries, we use the Chain of Density prompt that incorporates each video's metadata, based on the prompt proposed by Adams et al.~\cite{prompt}. 
On both datasets, our model is optimized by the Adam optimizer, with a maximum training epoch of 100. The threshold $\delta$ defined in Eq.~(\ref{eq:diversity}), to quantify the diversity of the input video is established at 0.35. The number of videos with $D$ below 0.35 is 14 for SumMe and 3 for TVSum.
To convert frame-level scores into scene-level scores, we segment the video into scenes as previous works~\cite{vsum,csnet,,graph,logcosh,CASUM,RSSUM}. First, we extract 1024-dimensional features from the pool5 layer of GoogLeNet~\cite{googlenet} pre-trained on ImageNet~\cite{imagenet} for the video frames. Using these frame-level features, the video is segmented into several scenes using the KTS algorithm~\cite{kts}. After calculating frame-level importance scores, these scores are aggregated into scene-level importance scores by averaging the scores within each scene. Finally, important scenes are selected by solving the knapsack problem, ensuring the video summary is 15\% of the original video's length, a common method for video summarization~\cite{vsum,csnet,sparsity,global,lstm,graph,logcosh,CASUM,RSSUM,M3SUM,a2summ,SSPVS,MSVA,scale,MTIDNet,iptnet,clipit,csta}.
More details about the experimental settings are provided in the supplementary material.

\subsection{Evaluation Metrics}
\noindent
When comparing our method with other state-of-the-art (SOTA) unsupervised video summarization methods, we omit the F-score, which measures the overlap between the predicted video summary and the reference summaries. This choice is based on research papers that suggest the F-score is unreliable in the video summarization task~\cite{rethinking,annotation}. The F-score is influenced by the common segmentation and segment selection process. Otani et al.~\cite{rethinking} demonstrated that even randomly selecting these pre-processed segments can achieve high F-scores. Also, when solving the knapsack problem, the model selects as many short scenes as possible instead of choosing a longer scene with higher scene-level score.
Therefore, we use two rank-based evaluations, Kendall's $\tau$~\cite{tau} and Spearman's $\rho$~\cite{rho}, proposed in~\cite{rethinking}, as evaluation metrics. 
In these metrics, the predicted frame-level scores are compared with the scores annotated by humans, which are independent of the segmentation and segment selection process.

\subsection{Results}

\begin{table}[t]
\centering
\caption{Comparison with SOTA unsupervised learning methods on the SumMe and TVSum datasets using Kendall's $\tau$ and Spearman's $\rho$ metrics. The \textbf{bolded} and \underline{underlined} items represent the best and second-best results.}
\label{tab:result}
\begin{tabular}{@{}lcccc@{}}
\toprule
\multirow{2}{*}{Methods}& \multicolumn{2}{c}{SumMe} & \multicolumn{2}{c}{TVSum} \\ 
\cmidrule(lr){2-3} \cmidrule(lr){4-5} 
 &  $\tau \uparrow$ & $\rho \uparrow$  & $\tau \uparrow$ & $\rho \uparrow$ \\ 
\midrule
Random~\cite{rethinking} & 0.000 & 0.000 &  0.000 & 0.000 \\
Human~\cite{rethinking} &  0.205 & 0.213  &  0.177 & 0.204 \\
\midrule
DRDSN~\cite{vsum} & 0.047 & 0.048  & 0.020 & 0.026\\
CSNet~\cite{csnet}  & - & - & 0.025 & 0.034 \\
$\text{RSGN}_u$~\cite{graph}  & \underline{0.071} & 0.073  & 0.048 & 0.052\\
DSAVS~\cite{logcosh} & - & - & 0.080 & 0.087\\
CASUM~\cite{CASUM} &  0.063& \underline{0.084} & \textbf{0.160} & \textbf{0.210}\\
RSSUM~\cite{RSSUM} & 0.007& 0.015 &  0.080& 0.106\\
\midrule
\textbf{Ours} & \textbf{0.102} &\textbf{0.138} & \underline{0.133} & \underline{0.174}\\ 
\bottomrule
\end{tabular}
\end{table}
We compared our proposed method with the SOTA unsupervised learning video summarization methods on the SumMe and TVSum datasets. The results are shown in Table~\ref{tab:result}, and our method achieves the best performances on SumMe and the second-best results on TVSum. CASUM shows the best performance on TVSum but it uses different $\epsilon$ values in Eq.~(\ref{regular}) for videos within the same dataset, which is significantly different from others that use the same hyperparameters for all videos within each dataset.
Moreover, our proposed method has another advantage in terms of computational cost. DSAVS and CASUM utilize Query-Key-Value attention mechanisms, and RSSUM utilizes multi-head attention mechanisms~\cite{transformer,fasttransformer}, which have a $\mathcal{O}(d \cdot n^2)$ computational complexity for $n$ frames and hidden dimension $d$, growing quadratically with the number of frames $n$. In contrast, ours
has a $\mathcal{O}(d \cdot n)$ computational complexity. This scales linearly with the video length, making it more efficient for processing longer videos.

\subsection{Model Analysis}
\begin{table*}[t]
\centering
\caption{The results of different self-supervised loss functions using  Kendall's $\tau$ and Spearman's $\rho$ metrics.}
\label{tab:loss}
\begin{tabular}{@{}llcccccc@{}}
\toprule
\multirow{2}{*}{Method}& \multirow{2}{*}{Loss}& \multicolumn{2}{c}{SumMe} & \multicolumn{2}{c}{TVSum} \\ 
\cmidrule(lr){3-4} \cmidrule(lr){5-6} 
 &  & $\tau \uparrow$ & $\rho \uparrow$   & $\tau \uparrow$ & $\rho \uparrow$ \\ 
\midrule
Improved margin ranking loss only&$\mathcal{L}_m$  & -0.015 & -0.019 &  \underline{0.121} & \underline{0.159} \\
Sparsity loss only~\cite{sparsity}&$\mathcal{L}_s$&  \underline{0.059} & \underline{0.080}  &  0.032 & 0.043 \\
AWL~\cite{awl}&$\frac{1}{2\sigma _1 ^2}\mathcal{L}_m + \frac{1}{2\sigma _2 ^2}\mathcal{L}_s + \text{log}\sigma _1 + \text{log}\sigma _2$& 0.037 &0.051 & 0.024 & 0.031\\ 
\midrule
\textbf{PDL (Ours)}&$\mathcal{L}_m + \lambda \mathcal{L}_s$&  \textbf{0.102} &\textbf{0.138} & \textbf{0.133} & \textbf{0.174}\\ 
\bottomrule
\end{tabular}
\end{table*}
Our proposed PDL is comprised of two components: the improved margin ranking loss (${L}_{m}$) and regularization loss (${L}_{s}$)~\cite{sparsity}. The effectiveness of our PDL is demonstrated in Table~\ref{tab:loss}, where it consistently achieves the highest scores on both datasets. In our analysis, we assess the impact of each loss individually
and explore the combination of loss functions in multi-task learning frameworks, which can learn relative weighting automatically from the data~\cite{awl}. 
In Table~\ref{tab:loss}, $\sigma_1$ and $\sigma_2$ act as observation noise scalers. 
For a fair comparison, we use the same experimental settings.

The results suggest that using only the improved margin ranking loss effectively optimizes the model for the TVSum dataset, which features numerous scene changes. Conversely, using only regularization loss is more effective for the SumMe dataset than the improved margin ranking loss alone, as it helps maximize the diversity in the generated summaries. 
The Automatic Weighted Loss (AWL)~\cite{awl} fails to account for the relative contributions and importance of each task in both datasets. 
On the other hand, our proposed PDL, which uses balanced parameters that take into account the diversity of the video, allows the model to dynamically prioritize between tasks, resulting in the most robust and effective model.
Therefore, we not only demonstrate the superior capabilities of our PDL but also highlight the significance of a tailored approach to loss function formulation that specifically considers the diversity of the video in the video summarization tasks.

\subsection{Visualization}
\begin{figure*}[t]
	\centering
	\subfloat[AWL~\cite{awl} ( $\tau$=-0.005, $\rho$=-0.007)]{\includegraphics[width=0.42\textwidth]{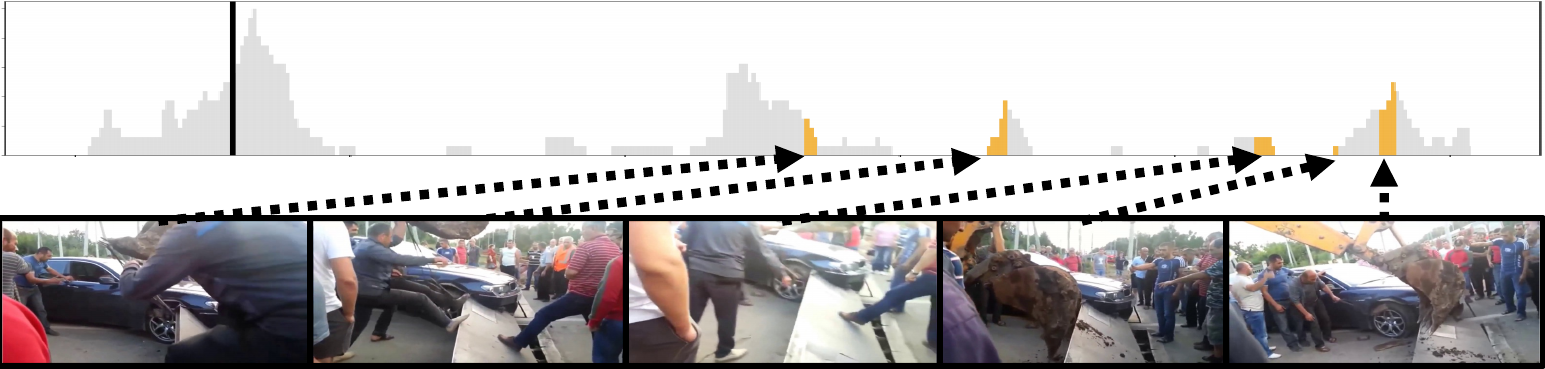}}\qquad \qquad
	\subfloat[\textbf{PDL (Ours)} ( $\tau$=0.164, $\rho$=0.236)]{\includegraphics[width=0.42\textwidth]{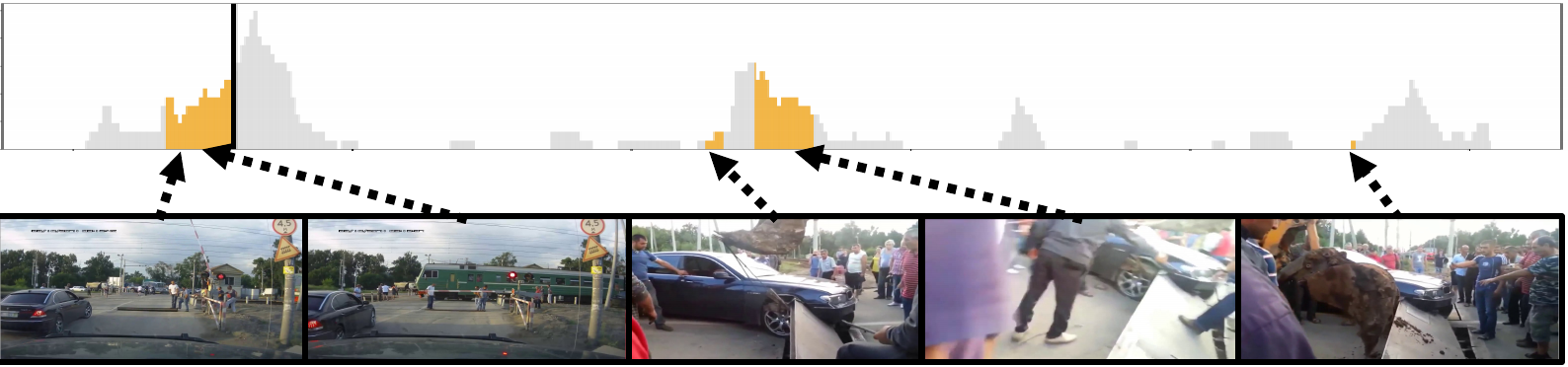}}\\
	\caption{Visualization results of the summarized video 6 in SumMe (``Car railcrossing'') generated by our models using different self-supervised loss functions. $D$ score of the video is 0.383. The light-gray bars in the figure represent the ground truth importance scores, while the orange areas indicate the parts selected by the model. The x-axis represents the frame index. The black vertical lines in the figure represent significant content changes within this video, with details documented in this section. The five images below are the representative frames selected as the video summary.}
	\label{fig:example_car}
\end{figure*}
\begin{figure*}[t]
	\centering
	\subfloat[AWL~\cite{awl} ( $\tau$=0.027, $\rho$=0.035)]{\includegraphics[width=0.42\textwidth]{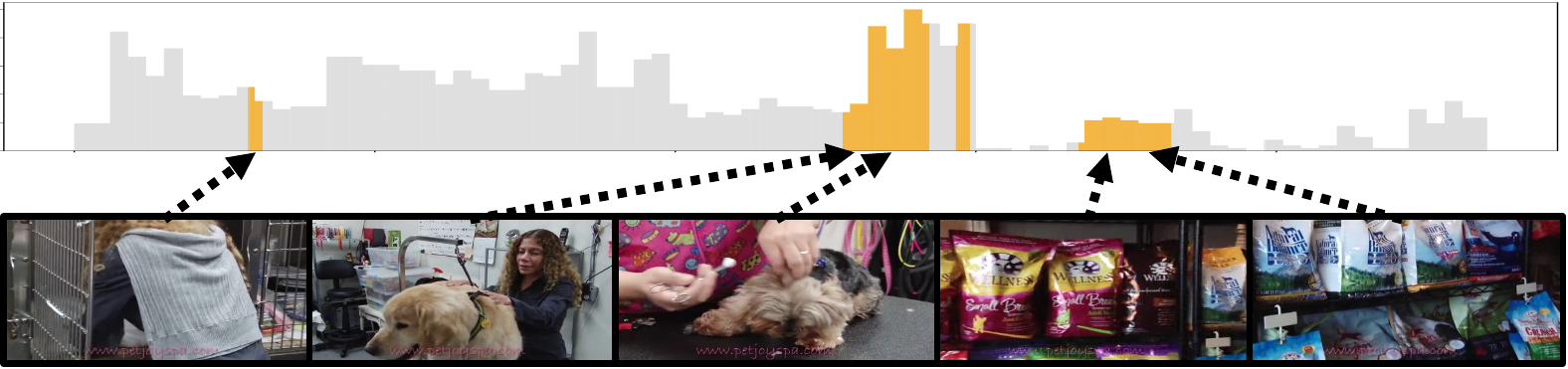}}\qquad \qquad
	\subfloat[\textbf{PDL (Ours)} ( $\tau$=0.167, $\rho$=0.227)]{\includegraphics[width=0.42\textwidth]{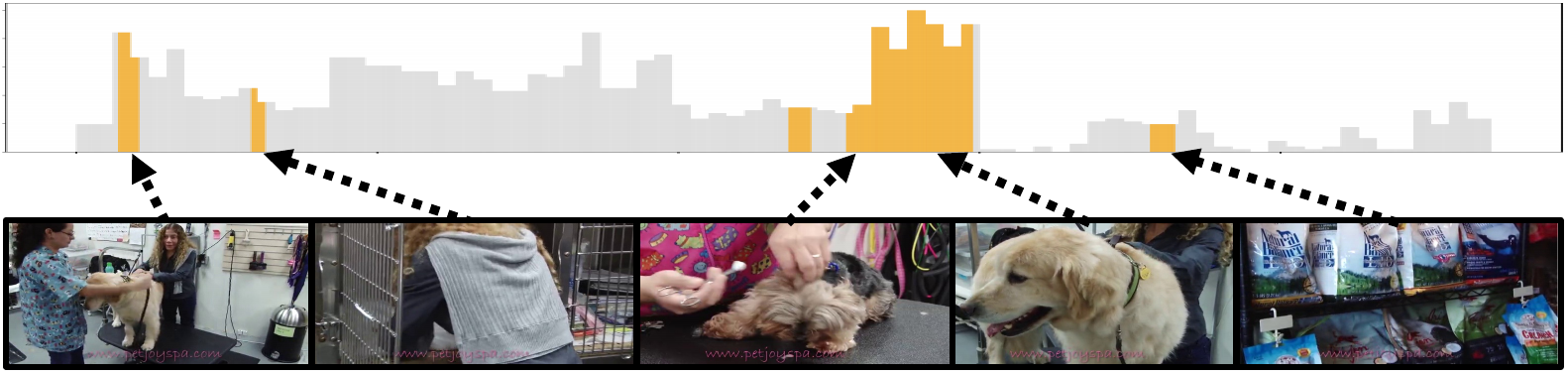}}\\
	\caption{Visualization results of the summarized video 11 in TVSum (``Pet Joy Spa Grooming Services - Brentwood, CA'') generated by our models using different self-supervised loss function. $D$ score of the video is 0.287.} 
	\label{fig:dog}
\end{figure*}
We provide the visualization results of video summaries generated by our proposed method, alongside comparisons with a model utilizing the multi-task uncertainty weighting approach proposed by Kendall et al~\cite{awl}. We specifically focus on video 6 from SumMe dataset and video 11 from TVSum dataset. The results are shown in Figure~\ref{fig:example_car} and Figure~\ref{fig:dog}.

The original video 6 in SumMe depicts a vehicle encountering an unexpected obstacle while crossing a railroad track, resulting in a collision. Following the incident, the video showcases a collaborative effort involving a backhoe loader and individuals working together to assist and recover the vehicle. In Figure ~\ref{fig:example_car}, the black vertical line delineates scenes before and after the collision, which divides the video content into two parts. The analysis of the result shown in Figure~\ref{fig:example_car}(a) reveals that the comparative method only includes scenes after the collision in the summarized video. In contrast, our proposed method shown in Figure~\ref{fig:example_car}(b) successfully captures scenes both before and after the collision, demonstrating its ability to encapsulate the entire storyline. This distinction underscores the effectiveness of our approach in providing a more comprehensive and contextually rich video summary.

The original video 11 in TVSum revolves around the facilities of a pet spa shop. Figure~\ref{fig:dog} shows that despite the minimal scene changes within the original video, the video summary generated by our proposed method is more diverse, successfully picking up qualitatively important parts from a wide range of content. 

Moreover, 
our proposed method demonstrates significantly higher $\tau$ and $\rho$, showcasing its utility and effectiveness in creating meaningful summaries. 
This indicates the capability to identify and include key moments, ensuring a comprehensive and engaging summary.

\subsection{Personalized Video Summarization}
\begin{figure*}[t]
    \centering
\includegraphics[width=0.41\linewidth]{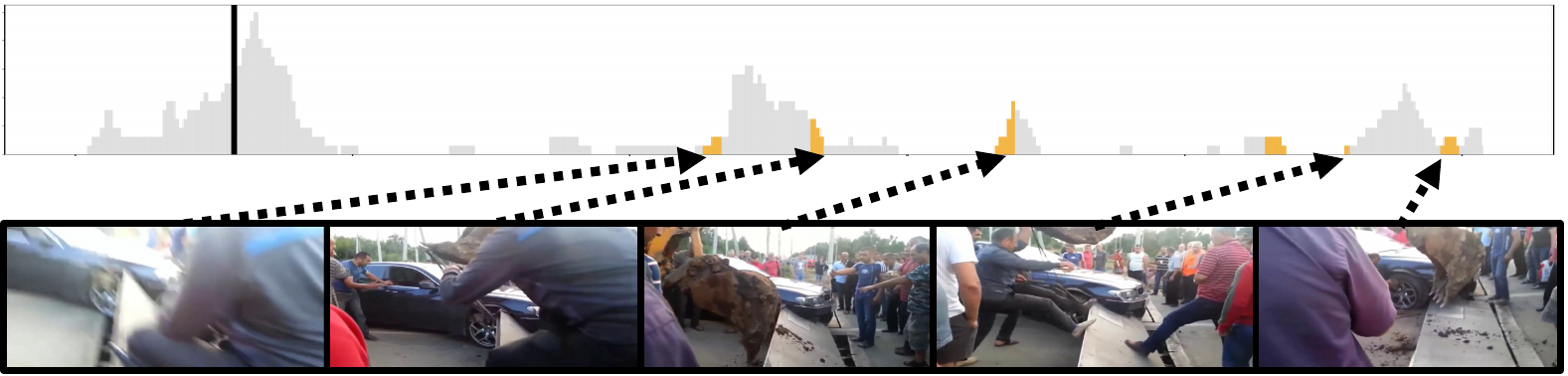}
    \caption{Visualization result of the personalized summarized video 6 in SumMe (``Car railcrossing'') generated by our model, where our model was guided by the LLMs to generate text summaries with a focus on the car after the accident.}
    \label{fig:personal_car}
\end{figure*}

We generate video summaries by calculating the similarity between the captions and the text summary produced by LLMs. This process allows us to influence the resulting video summary by adjusting the generated text summary according to user specifications. By incorporating user queries as prompts into the LLMs, we can flexibly control the content and focus of the text summary, inherently affecting the diversity and details emphasized in the video summary. 
The idea of leveraging LLMs to allow users to specify what they want to see in the video bridges the semantic gap, making the framework flexible enough to summarize videos across various domains and effectively achieve personalization.

For example, if you want to view footage of the car following an accident in the video 6 in SumMe, you can prompt the LLMs by stating, ``I would like to watch the video that focuses on the car after the accident.'' Consequently, the model generates the video shown in Figure~\ref{fig:personal_car}. In the generated video, a summarized version is generated that, as requested, focuses on the aftermath of the accident.
The prompt design and more examples using the Mr.HiSum dataset~\cite{mrhisum} are provided in the supplementary material.

\subsection{Limitations}
The limitation of our approach is that our model does not account for the temporal dependencies between input captions although LLMs consider temporal sequences when generating text summaries. Additionally, the captions generated by the image captioning model do not always perfectly describe the frames. We will address these issues in our future work.

\section{Conclusions}
In this paper, we propose a novel LLM-guided self-supervised video summarization framework. Our method eliminates the need for extensive data annotation and reduces subjectivity. We achieve frame-level scoring in the text semantic space. Additionally, we systematically analyze the characteristics of the datasets and mathematically define the diversity of the video. Subsequently, we construct a novel PDL function to create a more robust model tailored to the diversity of the video. The experimental results suggest that our proposed method achieves SOTA performance on the SumMe dataset and the second-best results on the TVSum dataset,
demonstrating the effectiveness of our approach. Additionally, our proposed framework flexibly enables the creation of personalized and customizable summaries tailored to the user's objectives by allowing users to direct the generation of text summaries by LLMs. This paper paves a new way for video summarization and is crucial for real-world scenarios where the video text description is not always available. We hope our framework will inspire further advancements in the field of video summarization.

\begin{acks}
This work is partially financially supported by the Beyond AI project of The University of Tokyo.  
\end{acks}

\clearpage
\appendix
This supplementary material provides the implementation details and further details of the personalized video summarization.
\noindent
\section{Implementation Details}
\subsection{Comparison of caption diversity}
We generate individual captions from downsampled frames using an image captioning model. One main reason for using an image captioning model instead of a video captioning model for caption generation is that using an image captioning model allows for more accurate assessment of the content of each frame and calculation of frame-level scores. 
Regarding the prompt, we compare three prompts: ``a scene of'', ``a frame of'', and ``a photo of'', using two image captioning models: the Generative Image-to-text transformer (GIT)~\cite{git} and the Bootstrapping Language-Image Pre-training with frozen unimodal models 2 (BLIP-2)~\cite{blip2}. The average percentages of generated unique captions within each video are shown in Table~\ref{tab:caption}.
Comparing the three prompts, ``a photo of'' results in the highest uniqueness in captions across both datasets and both image captioning models. When comparing the two image captioning models, GIT~\cite{git} demonstrates higher diversity compared to BLIP-2~\cite{blip2}, indicating minimal duplication and a lower occurrence of repetitive expressions. Therefore, we finally use GIT as the image captioning model and ``a photo of'' as the prompt to generate descriptive captions from individual downsampled video frames, as this combination results in more diverse captions. This diversity allows for a more accurate capture of the dynamic actions within the video, providing a richer and more detailed representation of the content.

\subsection{Experiment Details}
In the experiment investigating the contributions of the regularization loss in Section~3.3 of the main draft, we train our model with a learning rate of $1 \times10^{-5}$, and set the margin $m$ in Eq.~(3) to 0.15 for both the SumMe and TVSum datasets. 
For the general video summarization in Section~4.2 of the main draft, we set the learning rate of $5 \times10^{-5}$ and $1 \times10^{-5}$, the margin $m$ in Eq.~(3) to 0.11 and 0.06 for the SumMe and TVSum datasets, respectively. 

\subsection{Prompt for text summary generation}

\noindent\textbf{Prompt for generating personalized text summary.}
In section~4.7 of the main draft, we mentioned that our proposed framework allows for the customization of the generated text summary according to the user query.
Prompt~\ref{personalprompt} shows the detailed process of generating a personalized text summary. Specifically, we input the individually generated captions into [CAPTIONS] in chronological order, and the user query is fed into [USER QUERY].
\noindent
\begin{minipage}{\linewidth}
\begin{lstlisting}[label=personalprompt,caption=Our proposed prompt to generate text summary according to the user query.]
You are an expert in video summarization.

In this task, you will create concise and personalized summaries from captions that have been created from video frames by an image captioning model. 

Focus on extracting and highlighting only the elements of the captions that are directly relevant to the user's specific interests.

Your goal is to emphasize the most significant details pertinent to the query while omitting any information that is not relevant, ensuring that the summary is insightful and precisely tailored to the user's needs.

User query : """[USER QUERY]""""

Captions: """[CAPTIONS]"""

The final summary should be delivered in JSON format in a single line (~80 words), perfectly encapsulating the user's query with accuracy and relevance, and avoiding any extraneous content.

\end{lstlisting}
\end{minipage}
\setcounter{lstlisting}{0}
\renewcommand{\lstlistingname}{Text Summary}

\begin{table}[t]
\centering
\caption{Comparison of the average percentages of generated unique captions on the SumMe and TVSum datasets. The~\textbf{bolded} items represent the best results.}
\label{tab:caption}
\begin{tabular}{@{}lcccc@{}}
\toprule
\multirow{2}{*}{Prompt} & \multicolumn{2}{c}{SumMe~\cite{SumMe}} & \multicolumn{2}{c}{TVSum~\cite{TVSum}} \\ 
\cmidrule(lr){2-3} \cmidrule(lr){4-5} 
 & BLIP-2~\cite{blip2} & GIT~\cite{git} & BLIP-2~\cite{blip2} & GIT~\cite{git} \\ 
\midrule
``a scene of'' & 28 & \textbf{69}&  36 & \textbf{76} \\
``a frame of'' & 28 & 68 &  36 & \textbf{76} \\
``a photo of'' & \textbf{29} & \textbf{69} &  \textbf{37} & \textbf{76} \\
\bottomrule
\end{tabular}
\end{table}
\subsection{Examples for generated text summary}
In this section, we present examples of both general and personalized text summaries generated by GPT-4. Specifically, we use video~6 (``Car railcrossing'') from SumMe dataset. This video captures the entire sequence of a car crashing into an obstacle while crossing a railroad. Initially, while the railroad crossing is closed, cars and buses are visible in the scene. After the crossing opens and vehicles begin to move, a car encounters an unexpected obstacle on the railroad track, leading to a collision. Many people rush to the car to help. 

\noindent\textbf{Generated general text summary.}
A general text summary example, generated by GPT-4 using the Chain of Density prompt, is presented in Text Summary~\ref{car_summary}. It describes the scene in which many vehicles are visible while the railroad crossing is closed, and the crowd that gathers following the accident. This text summary captures both the moments before and after the accident, making it a suitable summary for the video.

\noindent
\begin{minipage}{\linewidth}
\begin{lstlisting}[label=car_summary,caption=Generated text summary for video 6 in SumMe (``Car railcrossing'').]
A green train, truck, and bus are halted at a railway crossing with a red and white barrier, stop lights, and 'train stop' signs. People, including those in white shirts, a black car on a muddy road, and a man in a blue shirt on a ramp, contribute to the scene's complexity. The presence of 'no parking' signs and the train's repeated stops at red lights highlight the strict control at the crossing.
\end{lstlisting}
\end{minipage}

\noindent\textbf{Generated personalized text summary according to the user query.}
Text Summary~\ref{car_personal} is generated based on the user query, ``I would like to watch the video that focuses on the car after the accident.'' The generated text summary captures only the car's condition after the accident and the gathering of people, completely omitting the state of the railroad crossing being closed. This example indicates that the generated text summary was successfully created in accordance with the user query.

\noindent
\begin{minipage}{\linewidth}
\begin{lstlisting}[label=car_personal,caption=Generated personalized text summary for video 6 in SumMe (``Car railcrossing'') with a focus on the car after the accident.]
Videos of interest primarily feature aftermath scenes of car accidents, highlighting damaged vehicles and their surroundings. Key visuals include a car smashed into a wall, another into a trailer, and various instances of cars with significant damage such as broken bumpers, windows, and large objects impacting them. Notably, a car is shown with a large rock on its front, another towed away, and scenes with a crowd gathered around, indicating the aftermath of accidents.
\end{lstlisting}
\end{minipage}

\section{Personalized video summarization}
\subsection{Dataset}
In this section, we present more details of the personalized video summarization using the videos in the Mr.HiSum~\cite{mrhisum} dataset. Mr.HiSum consists of YouTube videos within the YouTube-8M dataset~\cite{youtube8m}, specifically those that have highlight labels available on YouTube.

\subsection{Implementation Details}
The videos in the Mr.HiSum dataset are downsampled into one frame per second (fps) and scene boundaries are obtained using the KTS algorithm~\cite{kts}. Therefore, we downsample the video into one fps. The other experimental settings are the same as Section~4.7 of the main draft. 
We specifically use the movie trailers titled ``Restless | Official Trailer HD (2011)''~\footnote{\url{https://www.youtube.com/watch?v=pzUKbPAynxU}} and ``At Any Price Official Trailer''~\footnote{\url{https://www.youtube.com/watch?v=P_L35FyXqTA}} in the Mr.HiSum dataset. 
The original videos, named ``Restless\_original.mp4'' and ``At\_Any\_Price\_original.mp4'', are available in the folder linked on Google Drive~\footnote{\url{https://drive.google.com/drive/folders/1HPibyBmmEqWQGAeMJaRR2nJ0vbAY7nBs?usp=sharing}\label{video}}. 
For each video, we use two different user queries, providing two personalized video summaries per video.
The generated personalized videos are also provided as MP4 files~\footref{video}.

\subsection{Evaluation Metrics}
In order to quantitatively evaluate the generated personalized video summaries, we use recall and precision. Specifically, given the frames in the generated personalized video summary $\rm{P}$ and the ground truth frames that correspond to the user query $\rm{Q}$, the precision $\rm{p}$ and the recall $\rm{r}$ are defined as follows:
\begin{equation}
    \rm{p} = \frac{{\rm{P}}\cap {\rm{Q}}}{\text{Len}({\rm{P}})}, \quad
    {\rm{r}} = \frac{{\rm{P}}\cap {\rm{Q}}}{\text{Len}(\rm{Q})}. \quad
\end{equation}

\subsection{Generated text summaries}
The original video titled ``Restless | Official Trailer HD (2011)''  depicts a romantic story between a man and a woman. The man meets the woman by chance at a funeral, and as they get to know each other, their love gradually grows. 
Text Summary~\ref{scenery} and~\ref{conversation} show the results for the user queries ``I would like to watch a video that focuses on the background scenery and landscapes.'' and ``I would like to watch a video that focuses on the conversations and interactions between characters.'', respectively.

\noindent
\begin{minipage}{\linewidth}
\begin{lstlisting}[label=scenery,caption=Generated personalized text summary for ``Restless | Official Trailer HD (2011)'' with a focus on the background scenery and landscapes.]
The video features diverse backgrounds, including a man and a woman standing by a river with a red flag, highlighting a picturesque river scene. Two people sit on a bench next to a lake under an orange sky, offering a serene lakeside view. A couple sits on a bench by a lake, looking at trees and the water, emphasizing peaceful lake scenery. These scenes collectively showcase the video's focus on beautiful landscapes and tranquil outdoor settings.
\end{lstlisting}
\end{minipage}

\noindent
\begin{minipage}{\linewidth}
\begin{lstlisting}[label=conversation,caption=Generated personalized text summary for ``Restless | Official Trailer HD (2011)'' with a focus on the conversations and interactions between characters.]
The video features various interactions among characters, including a man and woman sitting at a dinner table, a couple discussing by a river, individuals conversing in a park, a woman talking to a man in a kitchen, and a couple kissing in the dark. It highlights personal and intimate moments, such as a boy talking to a girl, a couple walking and holding hands, and two people kissing on a sidewalk, emphasizing the focus on conversations and interactions between characters.
\end{lstlisting}
\end{minipage}

The original video titled ``At Any Price Official Trailer'' depicts the story of a famous farmer's son who defies his father's wishes and pursues his dream of becoming a professional race car driver instead of taking over the family business. In addition to the movie trailer, this video includes footage of a female speaker speaking at the beginning and end, which together make up roughly a quarter of the total runtime. 
Based on this content, we provide two types of user queries and evaluate whether the generated personalized video summaries correspond to the user queries.
Text Summary~\ref{carrelate} and~\ref{rural} show the results for the user queries ``I would like to watch a video that focuses on scenes related to cars.'' and ``I would like to watch a video that focuses on rural life and scenery.'', respectively.
\noindent
\begin{minipage}{\linewidth}
\begin{lstlisting}[label=carrelate,caption=Generated personalized text summary for ``At Any Price Official Trailer'' with a focus on the car related scenes.]
The video features various scenes related to cars, including a man and woman watching an orange race car, a woman waving to the crowd with a car on track, posters mentioning 'race car', scenes of a race car with the number 25, a group walking in front of a race car numbered 48, a car with 'clio' on the bottom in the middle of a race track, a couple kissing in a car with 'clio' visible, and a car driving down a dirt road kicking up dust."
\end{lstlisting}
\end{minipage}

\noindent
\begin{minipage}{\linewidth}
\begin{lstlisting}[label=rural,caption=Generated personalized text summary for ``At Any Price Official Trailer'' with a focus on the rural life and scenery.]
The video captures the essence of rural life and scenery, featuring a man in a field with a tractor and a large building in the background, a green tractor parked in front of a silo, and a farm with a silo and a red tractor. It also shows a man and woman looking out a window at a corn field, a red grain silo in front of a barn, and a car driving down a dirt road next to a field of corn, highlighting the agricultural and serene aspects of rural living."
\end{lstlisting}
\end{minipage}

\subsection{Visualizations and  Results}
\begin{figure*}[t]
	\centering
	\subfloat[Ground truth frames that are correspond to the user query. The frames highlighted in green are correctly selected by our model.]{\includegraphics[width=1.0\textwidth]{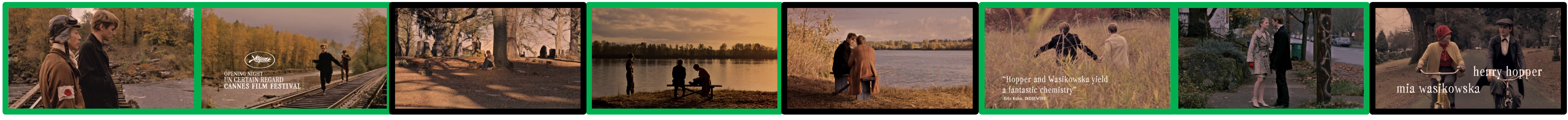}}\\
    \subfloat[The downsampled frames of our generated personalized video summary, (${\rm{p}}=0.714$). The numbers at the bottom left of each frame indicate the time it appears in the summarized video. The frames highlighted in green indicate those that are relevant to the user query, while the frames highlighted in red indicate those that are not.]{\includegraphics[width=1.0\textwidth]{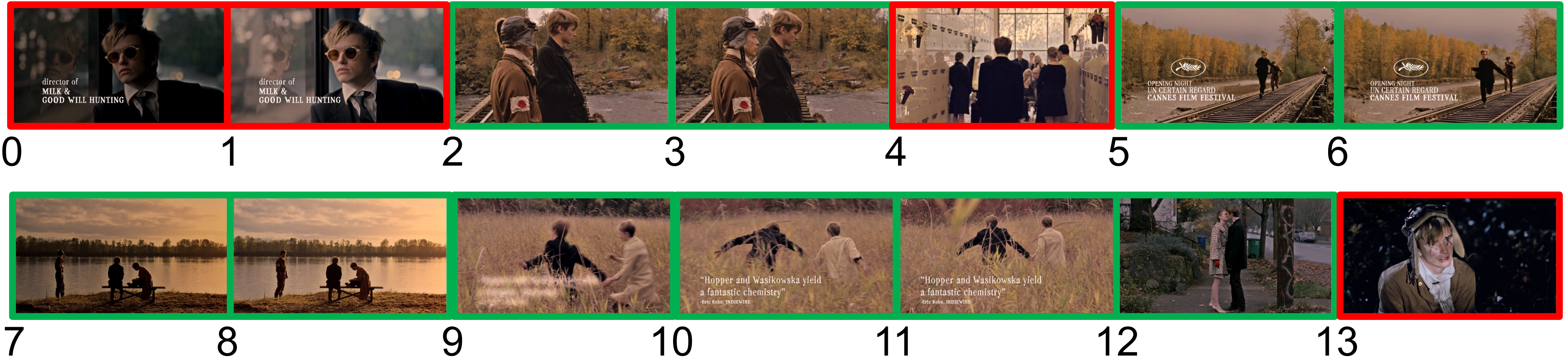}}\\
    \subfloat[Frames corresponding to the user query that our model selects. (${\rm{r}}=0.625$)]{\includegraphics[width=1.0\textwidth]{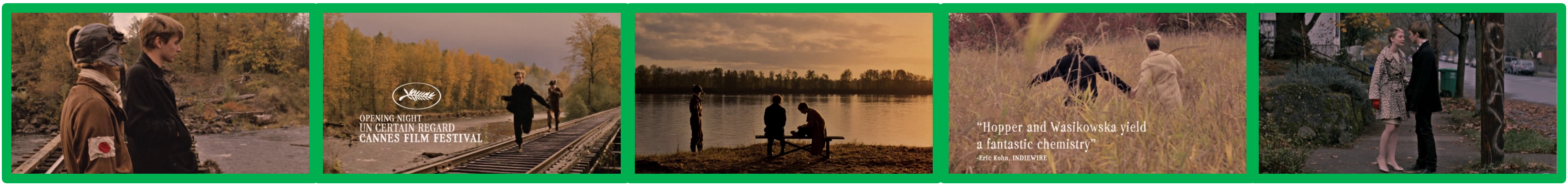}}\\
	\caption{Visualization of the video titled ``Restless | Official Trailer HD (2011)'' focusing on the background scenery and landscapes.}
	\label{fig:restless_scenery}
\end{figure*}
\begin{figure*}[t]
	\centering
	\subfloat[Ground truth frames that is correspond to the user query.]{\includegraphics[width=1.0\textwidth]{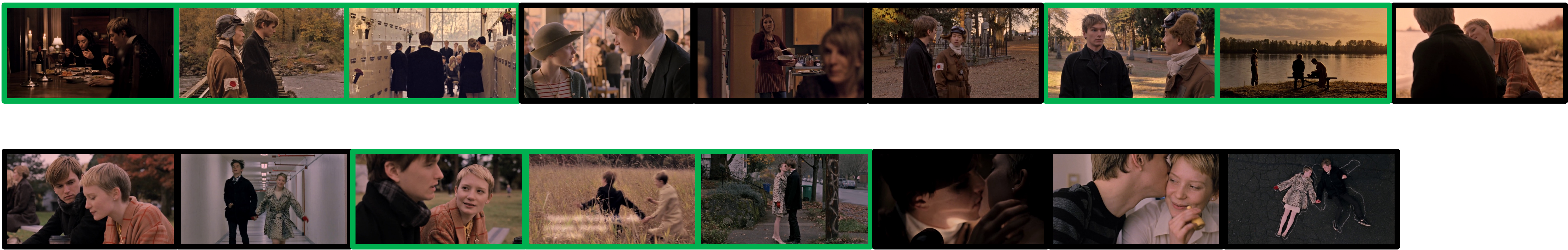}}\\
    \subfloat[The downsampled frames of our generated personalized video summary, (${\rm{p}}=0.929$).]{\includegraphics[width=1.0\textwidth]{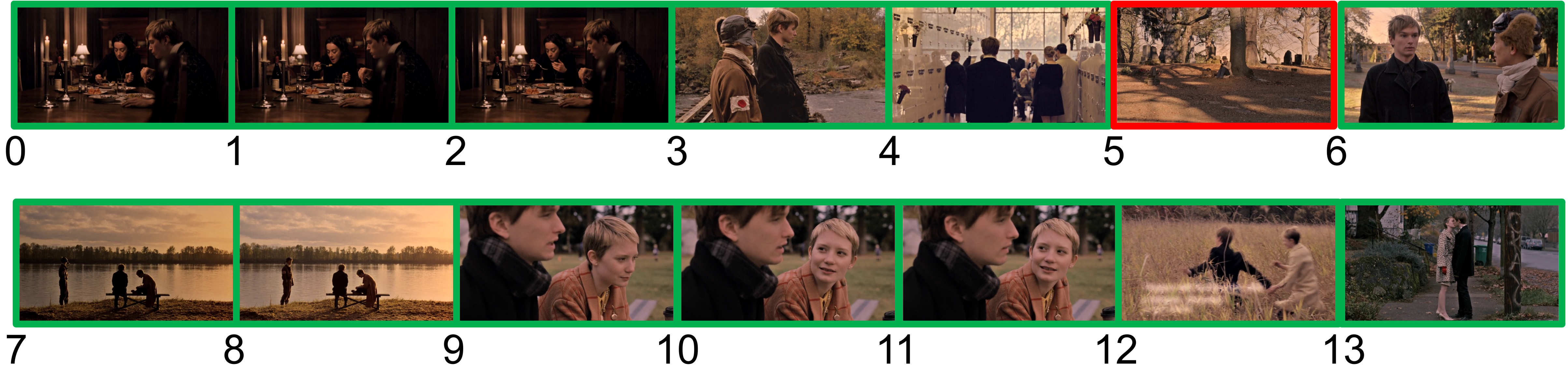}}\\
    \subfloat[Frames corresponding to the user query that our model selects. (${\rm{r}}=0.471$)]{\includegraphics[width=1.0\textwidth]{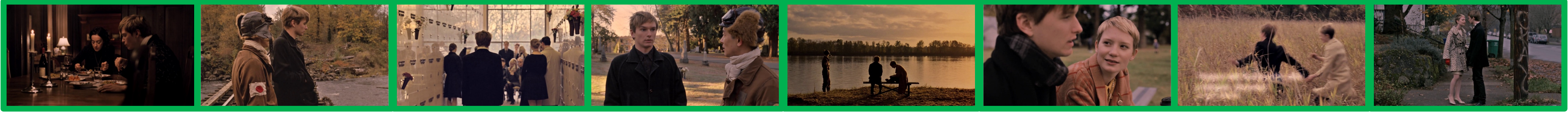}}\\
	\caption{Visualization of the video titled ``Restless | Official Trailer HD (2011)'' focusing on the conversations and interactions between characters.}
	\label{fig:restless_conversation}
\end{figure*}
\begin{figure*}[t]
	\centering
	\subfloat[Ground truth frames that is correspond to the user query.]{\includegraphics[width=1.0\textwidth]{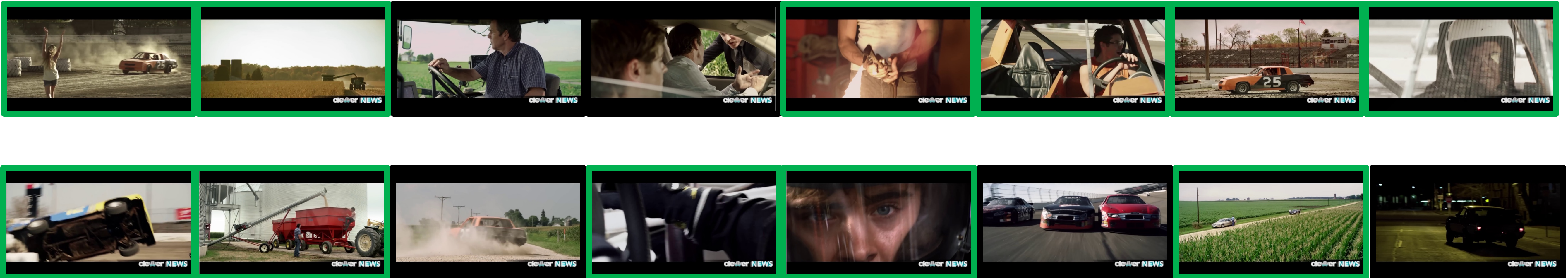}}\\
    \subfloat[The downsampled frames of our generated personalized video summary, (${\rm{p}}=0.682$).]{\includegraphics[width=1.0\textwidth]{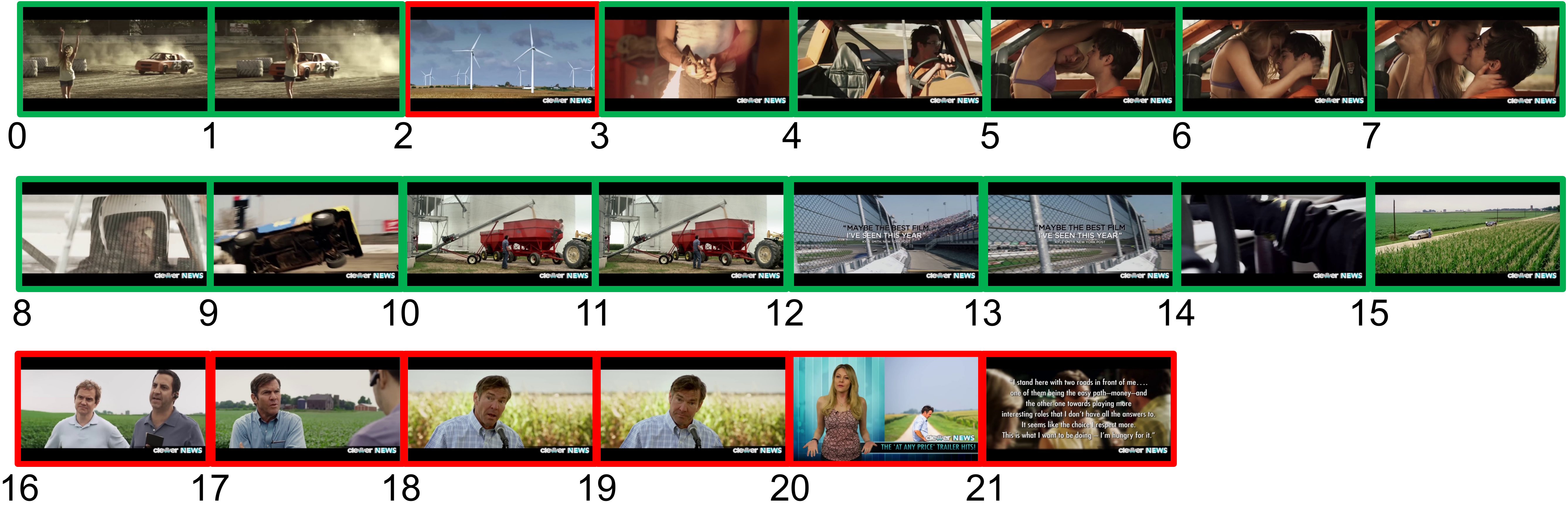}}\\
    \subfloat[Frames corresponding to the user query that our model selects, (${\rm{r}}=0.688$).]{\includegraphics[width=1.0\textwidth]{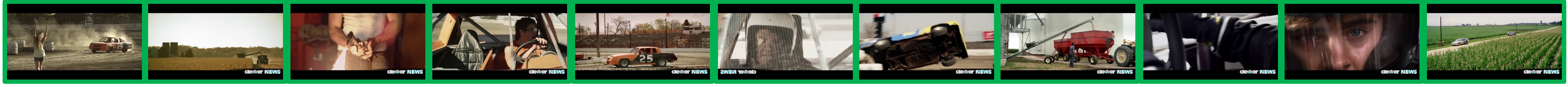}}\\
	\caption{Visualization of the video titled ``At Any Price Official Trailer'' focusing on the car related scenes.}
	\label{fig:at_car}
\end{figure*}
\begin{figure*}[t]
	\centering
	\subfloat[Ground truth frames that is correspond to the user query.]{\includegraphics[width=1.0\textwidth]{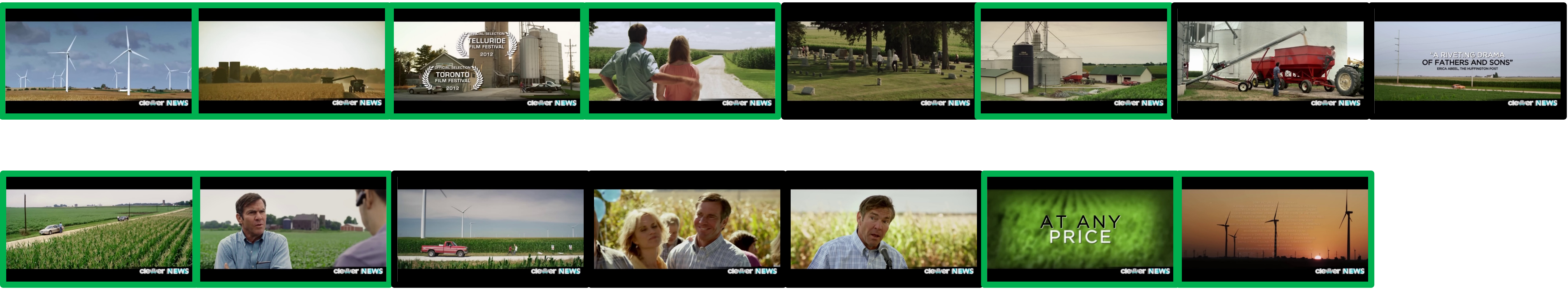}}\\
    \subfloat[The downsampled frames of our generated personalized video summary, (${\rm{p}}=0.652$).]{\includegraphics[width=1.0\textwidth]{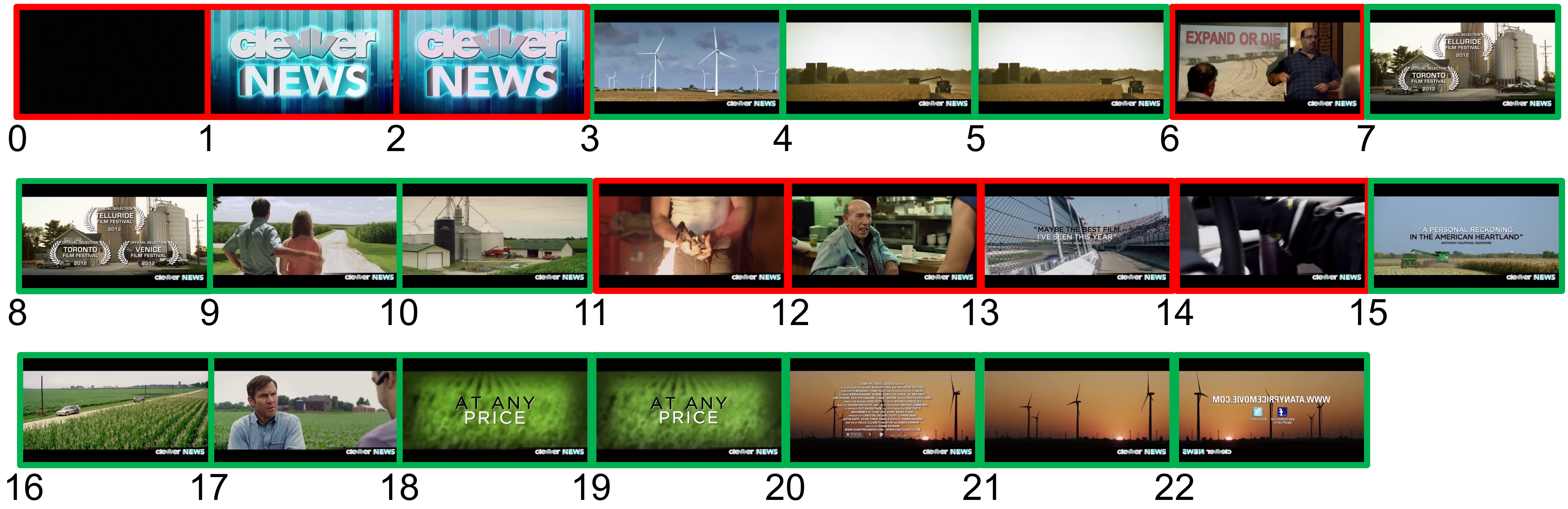}}\\
    \subfloat[Frames corresponding to the user query that our model selects, (${\rm{r}}=0.600$).]{\includegraphics[width=1.0\textwidth]{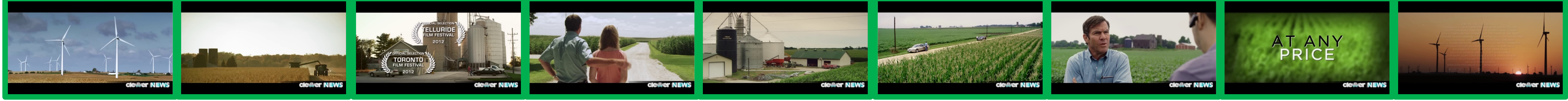}}\\
	\caption{Visualization of the video titled ``At Any Price Official Trailer'' focusing on the rural life and scenery.}
	\label{fig:at_rural}
\end{figure*}

Figure~\ref{fig:restless_scenery}(a),~\ref{fig:restless_conversation}(a),~\ref{fig:at_car}(a), and ~\ref{fig:at_rural}(a) show the ground truth frames we choose that are related to the user query. It should be noted that the video is downsampled to one fps. Additionally, if a frame corresponding to the user query exists within each scene, one representative frame is selected.
Also, Figure~\ref{fig:restless_scenery}(b),~\ref{fig:restless_conversation}(b),~\ref{fig:at_car}(b), and ~\ref{fig:at_rural}(b) show the downsampled frames of our generated personalized video, which has also been downsampled to one fps.
Additionally, Figure~\ref{fig:restless_scenery}(c),~\ref{fig:restless_conversation}(c),~\ref{fig:at_car}(c), and ~\ref{fig:at_rural}(c) show the correct frames generated by our model.

Figure~\ref{fig:restless_scenery} and~\ref{fig:restless_conversation} show the visualizations of the video titled ``Restless | Official Trailer HD (2011)'' when the user queries are ``I would like to watch a video that focuses on the background scenery and landscapes.'' and ``I would like to watch a video that focuses on the conversations and interactions between characters.'', respectively. The generated personalized videos, named ``Restless\_scenery.mp4'' and ``Restless\_conversation.mp4'', are available in the folder linked on Google Drive~\footref{video}.
Figure~\ref{fig:restless_scenery}(b) shows that out of the 14 frames in the generated personalized video, 10 frames correspond to the user query focusing on the background scenery and landscapes. Therefore, the precision is 0.714. 
Also, Figure~\ref{fig:restless_scenery}(c) shows that, out of the 8 ground truth frames, our model selects 5 frames. Therefore, the recall is 0.625. 
In the same way,  Figure~\ref{fig:restless_conversation} shows that out of the 14 frames in the generated personalized video, 13 frames correspond to the user query focusing on the conversations and interactions between characters (${\rm{p}}=0.929$). Additionally, out of 17 ground truth frames, our model selects 8 frames (${\rm{r}}=0.471$). 
These results indicate that, while some ground truth scenes are not selected due to the limitations of the summarized video length, the majority of the scenes within the generated personalized summaries correspond to the user query.
 

Figure~\ref{fig:at_car} and~\ref{fig:at_rural} show the visualizations of the video titled ``At Any Price Official Trailer'' for the user queries ``I would like to watch a video that focuses on scenes related to cars.'' and ``I would like to watch a video that focuses on rural life and scenery.'', respectively. The generated personalized videos, named ``At\_Any\_Price\_car.mp4'' and ``At\_Any\_Price\_rural.mp4'', are available in the Google Drive link~\footref{video}.
Figure~\ref{fig:at_car}(b) shows that out of the 22 frames in the generated personalized video, 15 frames correspond to the user query focusing on the car related scenes (${\rm{p}}=0.682$).
Also, Figure~\ref{fig:at_car}(c) shows that, out of the 16 ground truth frames, our model selects 11 frames (${\rm{r}}=0.688$). 
Similarly, Figure~\ref{fig:restless_conversation} shows that out of the 23 frames in the generated personalized video, 15 frames correspond to the user query focusing on the rural life and scenery (${\rm{p}}=0.652$). Additionally, out of 15 ground truth frames, our model selects 9 frames (${\rm{r}}=0.600$). 
These results demonstrate that the generated personalized video summaries successfully reflect distinctly different user queries, such as those related to cars and those depicting rural life and scenery, by creating personalized text summaries tailored to each user query.

\subsection{Limitations}
The limitation of our personalized video summarization approach is that we did not perform parameter searching for the Mr.HiSum dataset, so we believe further optimization of the parameter settings can ensure better results.
Additionally, we solve the knapsack problem to extract a subset of scenes. Among the scenes corresponding to the user query, some scenes corresponding to the user query are not selected due to their long length, despite having high scene-level scores. This issue arises from solving the knapsack problem to create the summary, as mentioned in Section~4.3 of the main draft.

\bibliographystyle{ACM-Reference-Format}
\bibliography{sample-base2}

\end{document}